\def\year{2019}\relax
\documentclass[letterpaper]{article} 
\usepackage{aaai19}  
\usepackage{times}  
\usepackage{helvet}  
\usepackage{courier}  
\usepackage{url}  
\usepackage{graphicx}  
\usepackage{amsmath}
\usepackage{amsfonts}
\usepackage{amssymb}
\usepackage{amsthm}
\usepackage{bbm}
\usepackage{mathtools}
\usepackage{blkarray}
\newtheorem{theorem}{Theorem}

\newtheorem*{corollary*}{Proof Sketch}
\newtheorem*{assumption*}{Assumption}

\usepackage[linesnumbered, algoruled]{algorithm2e}
\usepackage{subcaption}
\usepackage{booktabs}
\usepackage{subfiles}
\usepackage{color}
\begin{document}
%
\title{Robust Ordinal Embedding from Contaminated Relative Comparisons}
\author
{
		Ke Ma\textsuperscript{1,2}, Qianqian Xu\textsuperscript{3}, Xiaochun Cao\textsuperscript{1}\thanks{The corresponding authors.}\\
		\textsuperscript{1} State Key Laboratory of Information Security, Institute of Information Engineering, Chinese Academy of Sciences\\
		\textsuperscript{2} School of Cyber Security, University of Chinese Academy of Sciences\\
		\textsuperscript{3} Key Laboratory of Intelligent Information Processing, Institute of Computing Technology, Chinese Academy of Sciences\\
		\{make, caoxiaochun\}@iie.ac.cn, xuqianqian@ict.ac.cn\\
}
\maketitle
\begin{abstract}
Existing ordinal embedding methods usually follow a two-stage routine: outlier detection is first employed to pick out the inconsistent comparisons; then an embedding is learned from the clean data. However, learning in a multi-stage manner is well-known to suffer from sub-optimal solutions. In this paper, we propose a unified framework to jointly identify the contaminated comparisons and derive reliable embeddings. The merits of our method are three-fold: (1) By virtue of the proposed unified framework, the sub-optimality of traditional methods is largely alleviated; (2) The proposed method is aware of global inconsistency by minimizing a corresponding cost, while traditional methods only involve local inconsistency; (3) Instead of considering the nuclear norm heuristics, we adopt an exact solution for rank equality constraint. Our studies are supported by experiments with both simulated examples and real-world data. The proposed framework provides us a promising tool for robust ordinal embedding from the contaminated comparisons.
\end{abstract}

\section{Introduction}

The solutions to many tasks involve the similarity estimation of data samples, \textit{e.g.} clustering \cite{DBLP:journals/sac/Luxburg07}, classification \cite{DBLP:journals/jmlr/ChenGGRC09} and representation learning \cite{wei2018video}. Traditionally, the objective similarity measurement extracts the features of data points to calculate their distances in some space, while the Euclidean distance, cosine similarity, Hamming distance and Kullback-Leibler divergence are the favorite measurements. In real-world scenarios, the situation becomes complicated since the off-the-shelf similarity functions are unreliable. It is too difficult to customize a similarity function for the specific data, or simply unavailable of the features or attributes. Consequently, the subjective method is more appropriate when the objective criteria are ambiguous. Among various subjective approaches for similarity estimation, relative comparison is expected to yield more reliable results. Instead of evaluating the similarity on an absolute scale, the relative similarity only needs individuals to answer  a ``yes or no'' question as:
\textit{``Is the similarity between object $i$ and $j$ larger than the similarity between $l$ and $k$?''}
\noindent Then a point configuration is constructed to preserve the above constraints as much as possible. This task is known as the ordinal embedding. The problem first arises in the psychometric society \cite{Shepard1962a,Shepard1962b,Kruskal1964a,Kruskal1964b}. In recent years, it has gained increasingly attention. \cite{agarwal2007generalized,tamuz2011adaptiive,kevin2011active,vandermaaten2012stochastic,53e99af7b7602d97023851bf,Terada2014LocalOE,amid2015multiview,NIPS2016_6554,DBLP:conf/aaai/MaZXXCLY18}


However, the relative comparison approach leaves a cumbersome burden on participants with a large number of annotations. Due to its economical and scalable implementation, crowdsourcing platforms (\textit{e.g.}, MTurk, Innocentive, CrowdFlower, CrowdRank, and Allourideas) are always restored to annotate these relative comparisons. Nevertheless, crowdsourced relative comparisons are not without pitfalls $-$ the crowd is not all trustworthy \cite{DBLP:journals/tmm/XuHJYLY12,DBLP:conf/wsdm/ChenBCH13}. Since participants perform experiments without supervision on the Internet, when the testing time for a single participant lasts too long, the annotators may give untrustworthy feedbacks. Such unreliable labels bring great challenges to control the quality of crowsourced relative comparisons, let alone utilizing them in the subsequent tasks. Therefore, how to derive reliable embeddings from these contaminated data has become an urgent issue in the ordinal embedding research.


Traditional robust ordinal embedding methods are usually two-staged: (1) The outlier detection is employed to pick out the inconsistent comparisons. (2) Based on the cleaned data, the embedding is constructed to map items into a Euclidean space with a low dimension. Various methods have been developed in literature for outlier detection, such as majority voting \cite{DBLP:conf/nips/WelinderBBP10}, M-estimator \cite{huber2009robust}, Least Median of Squares (LMS) \cite{rousseeuw1984least}, S-estimators \cite{rousseeuw1984robust}, Least Trimmed Squares (LTS) \cite{rousseeuw2005robust}, and Thresholding based Iterative Procedure for Outlier Detection \cite{she2011outlier} etc. Among these studies, perhaps the most well-known one is majority voting. A large budget is allocated to obtain multiple annotations for each comparisons. These annotations are then aggregated so as to eliminate label noise \cite{McFee:2011:LMS:1953048.1953063}. However, the effectiveness of the majority voting strategy is often limited by the sparsity problem $-$ it is typically infeasible to have many annotators for each relative comparisons. Moreover, it has been found that when pairwise local rankings are integrated into a global ranking, it is possible to detect outliers that can cause global inconsistency and yet are locally consistent, \textit{i.e.}, supported by majority votes \cite{DBLP:journals/mp/JiangLYY11}. Worse, the existing ordinal embedding models, such as GNMDS \cite{agarwal2007generalized}, CKL \cite{tamuz2011adaptiive} and STE \cite{vandermaaten2012stochastic,NIPS2016_6554}, only adopt the classification scheme via predicting the labels of the relative comparisons to construct the embeddings. The generalization of the embeddings would be damaged grievously when the labels of the training samples are wrong. As a consequence, separating as two unrelated parts, the outlier detection and the classification-based embedding would only obtain the local optimal solutions individually.

In this paper we propose a unified approach to detect outliers in contaminated data and derive robust ordinal embedding simultaneously. Specifically, instead of detecting outliers locally and independently for each comparison, our method works globally in a sense that the global inconsistency is explicitly penalized by the loss function. This enables us to identify those outliers which would be locally consistent with the majority results but in fact lead to significant global ranking inconsistency. The proposed model considers a partially penalized LASSO problem in the semi-definite programming. An efficient algorithm is proposed to obtain the embedding with exact desired dimension. The experiments are carried out on synthetic and real-world datasets. The results demonstrate that our method outperforms the state-of-the-art alternatives.

\section{Robust Ordinal Embedding}

\subsection{Preliminaries}


Let $\mathcal{O}=\{\boldsymbol{o}_1,\dots,\boldsymbol{o}_n\}$ be a set of objects which need to obtain the embedding. There exists an unknown but fixed similarity function $\zeta:\ \mathcal{O}^2\rightarrow\mathbb{R}^+$ which assigns the similarity value $\zeta_{ij}$ to a pair of objects $(\boldsymbol{o}_i, \boldsymbol{o}_j)$. In this sense, the ranking of $\{\zeta_{ij}\},\ i,j\in[n]$ will produce a total order. However, without any prior knowledge, $\mathcal{O}$ and $\{\zeta_{ij}\}$ are both unknown. Therefore, we establish the form of side-information that will drive our ordinal embedding algorithm, that is, the relative similarity measurements collected from human labelers. 

Given a set of objects $\mathcal{O}$ and a set of annotators $\mathcal{U}$, a collection of relative similarity measurements can be written as 
\begin{equation}
	\label{set:quadruple}
	\begin{aligned}
		\mathcal{C}_{\mathcal{U}}=\left\{(i,j,l,k)_{u}\ 
		\begin{array}{|c}
		i,\ j,\ l,\ j\in[n],\ u\in\mathcal{U}\\ 
		i\neq j,\ l\neq k,\ (i,j)\neq(l,k)
		\end{array}
		\right\},
	\end{aligned}
\end{equation}
where a tuple $(i, j, l, k)_u$ is interpreted as ``worker $u$ annotates that $i$ and $j$ are more similar than $l$ and $k$''. (This measurement subsumes the triple-wise comparison situation when $i = l$.) The goal of ordinal embedding is to find an embedding $\boldsymbol{X}=\{\boldsymbol{x}_1,\dots,\boldsymbol{x}_n\}\in\mathbb{R}^{p\times n}$ such that
\begin{equation}
	\label{eq:ordinal_constraint}
	d(\boldsymbol{x}_i,\boldsymbol{x}_j)<d(\boldsymbol{x}_l,\boldsymbol{x}_k),\ \forall\ (i,j,l,k)_u\in\mathcal{C}_{\mathcal{U}},
\end{equation}
where $d:\mathbb{R}^p\times\mathbb{R}^p\rightarrow\mathbb{R}^+$ is a distance function of Euclidean space $\mathbb{R}^p$. As both $\mathcal{O}$ and $\zeta$ lose the explicit form in ordinal embedding problem, the squared Euclidean distance $d(\boldsymbol{x}_i,\boldsymbol{x}_j)=\|\boldsymbol{x}_i-\boldsymbol{x}_j\|^2_2:=d_{ij}$ is always adopted in \eqref{eq:ordinal_constraint}. We denote $\boldsymbol{D}=\{d_{ij}\}$ as the distance matrix of $\boldsymbol{X}$.

Despite the distance matrix $\boldsymbol{D}$ is directly related to the embedding $\boldsymbol{X}$, the squared Euclidean distance is a nonlinear function of $\boldsymbol{X}$. Here we introduce the Gram matrix of $\boldsymbol{X}$ and conduct the distance as a linear function of Gram matrix. It is known that there is a map between the distance matrix $\boldsymbol{D}$ and the Gram matrix $\boldsymbol{G}=\{g_{ij}\}=\boldsymbol{X}^\top\boldsymbol{X}$ as
\begin{subequations}
	\label{eq:Gram}
	\begin{align}
		 d_{ij} &=g_{ii}-2g_{ij}+g_{jj},\\
		 \boldsymbol{D} &=\textit{diag}(\boldsymbol{G})\cdot\boldsymbol{1}^\top-2\boldsymbol{G}+\boldsymbol{1}\cdot\textit{diag}(\boldsymbol{G})^\top,
	\end{align}
\end{subequations}
where $\textit{diag}(\boldsymbol{G})$ is the column vector composed of the diagonal entries of $\boldsymbol{G}$ and $\boldsymbol{1}$ is the $n$-dimension vector whose all entries equal to $1$.

To dissect the underlying geometrical structure of $\mathcal{C}_{\mathcal{U}}$, we represent $\mathcal{C}_{\mathcal{U}}$ as a directed graph $\mathcal{G}=\{\mathcal{V},\mathcal{E}\}$ over $\mathcal{O}^2$ and note $c_u$ as a ordered tuple $(i,j,l,k)_u$. Each vertex $v_{ij}\in\mathcal{V}\subseteq\mathcal{O}^2$ in the graph $\mathcal{G}$ corresponds to a pair $(\boldsymbol{o}_i, \boldsymbol{o}_j)\in\mathcal{O}^2$, and an edge $e^u_c\in\mathcal{E}$ from $v_{ij}$ to $v_{lk}$ corresponds to a relative similarity measurement labeled by worker $u\in\mathcal{U}$. The measurement between $(\boldsymbol{o}_i, \boldsymbol{o}_j)$ and $(\boldsymbol{o}_l, \boldsymbol{o}_k)$ will be labeled by different annotators and their answers to the same question would be inconsistent. It leads to the multiple edges with different directions between $v_{ij}$ and $v_{lk}$. Let $\boldsymbol{e}^{\mathcal{U}}_{c}=\{e^u_c, u\in\mathcal{U},\ c\in\mathcal{C}_{\mathcal{U}}\}$ be the multiple edge with each single edge $e^u_c$ has the same direction, and we assign an indicator $y^u_c$ on $e^u_c$ as $ y^u_c = 1$ if $e^u_c$ existed. It means that worker $u$ measures $(\boldsymbol{o}_i, \boldsymbol{o}_j)$ and $(\boldsymbol{o}_l, \boldsymbol{o}_k)$, then she/he gives an answer which supposes that $\zeta_{ij}>\zeta_{lk}$. Notice that $y^u_c$ is skew-symmetric, for each $u\in\mathcal{U}$, \textit{i.e.}, $y^u_c = - y^u_{\bar{c}}$ where $c=(i,j,l,k)$ and $\bar{c} = (l,k,i,j)$. Furthermore, all the comparisons, from $(\boldsymbol{o}_i, \boldsymbol{o}_j)$ to $(\boldsymbol{o}_l, \boldsymbol{o}_k)$, are then aggregated over all annotators who have cast a vote on the two pairs. The results are represented as the weight of $\boldsymbol{e}^{\mathcal{U}}_{c}$, the total number of annotations on $c$, 
\begin{equation}
	\label{eq:edge_weight}
	w_c = \underset{u\in\mathcal{U}}{\sum}[y^u_c=1],\ c\in\mathcal{C}_{\mathcal{U}},
\end{equation}
where $[\cdot]$ indicates the Iverson's bracket notation. We denote the whole edge weight $\boldsymbol{w} = \{w_c\}\in\mathbb{R}^{|\mathcal{E}|}$. 

\subsection{Existing Problems of Traditional Methods}

Interpreting $\mathcal{C}_{\mathcal{U}}$ as the comparison graph $\mathcal{G}$ with multiple-edge $\{\boldsymbol{e}^{\mathcal{U}}_c\}$ will help us to infer global structure properties of $\mathcal{C}_{\mathcal{U}}$. In an ideal case, we know the similarity function $\zeta$ explicitly, and the global ranking of $\{\zeta_{ij}\}$ will leads the comparisons on $\{(\boldsymbol{o}_i,\boldsymbol{o}_j)\}\in\mathcal{O}^2$ to be a partial order. Two facts will become immediately apparent: 1) $\mathcal{G}$ is acyclic, and 2) the votes received on each edge are unanimous, \textit{e.g.} $w_c\geq1$ and $w_{\bar{c}}=0$. However, it is always difficult to design such a function $\zeta$ to measure the similarity of objects in $\mathcal{O}$. That's why we need the wisdom of a crowd and measure the relative similarity by the comparisons $\mathcal{C}_{\mathcal{U}}$ from human beings. There always exists disagreement in $\mathcal{C}_{\mathcal{U}}$ as both $w_c>0$ and $w_{\bar{c}}>0$ would appear. Assuming these opposite annotations cannot be true concurrently, one of them will be the outlier which should not be considered in the ordinal embedding problem. The traditional methods require a pretreatment of $\mathcal{C}_{\mathcal{U}}$. It is known as the \textit{majority voting} method which treats one direction with larger weight as the majority and the other as the minority, then the latter one will be pruned. Obviously, \textit{majority voting} is a ``\textit{local}'' outlier detection method and it ignores the potential ``\textit{global}'' similarity ranking on $\mathcal{O}^2$. 

\begin{figure}[t]
	\centering
	\includegraphics[width = 0.45\textwidth]{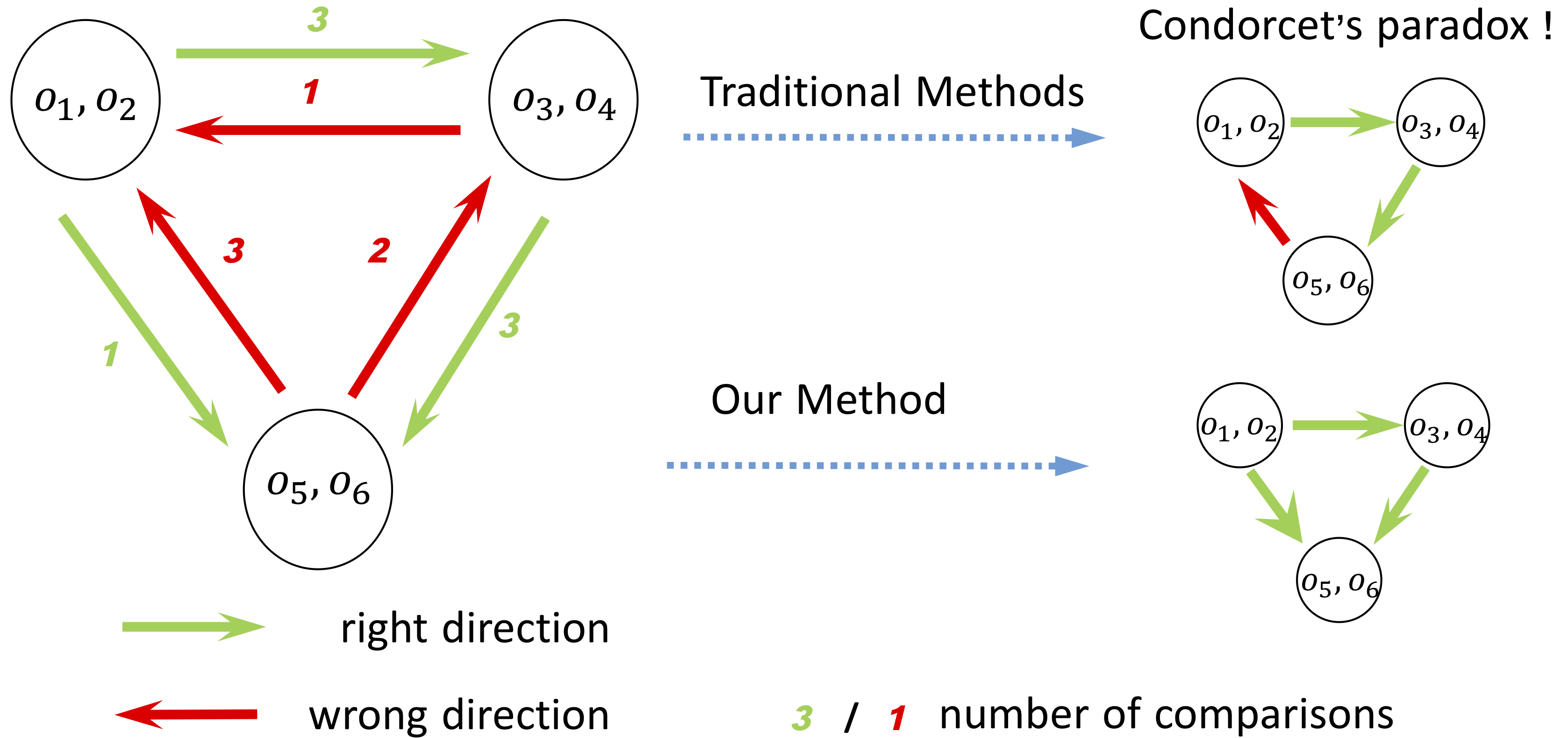}
	\caption{Traditional vs. Our methods in outlier detection. Green arrows/edges indicate correct annotations, while red arrows represent the outliers. The numbers indicate the number of votes received by each edge.}
	\label{fig:comparion_graph}
\end{figure}

To make the matter worse, this local treatment would keep the wrong annotations and remove the true labels (see Figure \ref{fig:comparion_graph}.) Suppose that $\zeta_{12}>\zeta_{34}>\zeta_{56}$, \textit{majority voting} will remove the edges from node $v_{12}$ to node $v_{56}$ as they are the minority versus the other direction edges. In particular, the \textit{majority voting} will introduce a cyclic comparison $v_{12} > v_{34} > v_{56} > v_{12}$ which is the well-known \textbf{Condorcet's paradox} \cite{gehrlein2006condorcet,DBLP:journals/mp/JiangLYY11}, and consequently $\mathcal{G}$ will contains cycles. In order to eliminate the cycles in $\mathcal{G}$, the maximum acyclic subgraph are adopted \cite{McFee:2011:LMS:1953048.1953063,vandermaaten2012stochastic} to replace $\mathcal{G}$. 

Summarizing the arguments, we have the following comments on the traditional methods. For one thing, these methods could not utilize the ordinal information $\mathcal{C}_{\mathcal{U}}$ properly by adopting \textit{majority voting} to outlier detection. For the other thing, an \textbf{NP-complete} problem, see maximum acyclic subgraph \cite{DBLP:books/fm/GareyJ79}, makes the whole process more complicated. These are the main motivations for us to propose a new framework which can detect the outlier and obtain the embedding simultaneously.

\subsection{Framework Formulation}

Given the comparison graph $\mathcal{G}$ with inconsistent multiple edge $\mathcal{E} = \{\boldsymbol{e}^{\mathcal{U}}_c, \boldsymbol{e}^{\mathcal{U}}_{\bar{c}}\ |\ c,\ \bar{c}\in\mathcal{C}_{\mathcal{U}}\}$, there are two goals:
\begin{itemize}
	\item Detecting the outliers in the edge set $\mathcal{E}$. To this end, we introduce a set of unknown variables $\boldsymbol{\gamma}=\{\gamma_c\}\in\mathbb{R}^{|\mathcal{E}|}$ where each variable $\gamma_c$ indicates whether the edge $\boldsymbol{e}^{\mathcal{U}}_{c}$ is an outlier or not. The outlier detection task in $\mathcal{C}_{\mathcal{U}}$ thus becomes the problem of estimating $\boldsymbol{\gamma}$ with $\mathcal{G}$.
	\item Obtaining an embedding $\boldsymbol{X}\in\mathbb{R}^{p\times n}$. Without prior knowledge, the squared Euclidean distance is selected as the dissimilarity measure, that is, given $\boldsymbol{e}^{\mathcal{U}}_c\in\mathcal{E}$ which represents the correct relative similarity measurement, we hope the embedding $\boldsymbol{X}$ satisfies 
	\begin{equation*}
		\|\boldsymbol{x}_i-\boldsymbol{x}_j\|^2_2 < \|\boldsymbol{x}_l-\boldsymbol{x}_k\|^2_2,\ c_u=(i,j,l,k)_u\in\mathcal{C}_{\mathcal{U}}.
	\end{equation*}
\end{itemize}
In contrast to the multi-staged methods, we propose to jointly learn the embedding $\boldsymbol{X}$ by $\mathcal{G}$ and remove outliers globally via $\boldsymbol{\gamma}$ to avoid finding the maximum acyclic subgraph of $\mathcal{G}$. For this purpose, the outlier indicator $\boldsymbol{\gamma}$ and the embedding $\boldsymbol{X}$ are estimated in a unified framework. Given an edge $\boldsymbol{e}^{\mathcal{U}}_c\in\mathcal{E}$, its corresponding direction indicator $y_c$ is modeled as
\begin{equation}
	\label{eq:linear_model}
	y_c = \|\boldsymbol{x}_i-\boldsymbol{x}_j\|^2_2-\|\boldsymbol{x}_l-\boldsymbol{x}_k\|^2_2+\gamma_c+\varepsilon_c, 
\end{equation}
where $\varepsilon_c\sim\mathcal{N}(0,\sigma^2)$ is the Gaussian noise with zero mean and a variance $\sigma$. The outlier indicator $\gamma_c$ is assumed to have a larger magnitude than $\sigma$. For multiple edge $\boldsymbol{e}^{\mathcal{U}}_c$, if they are not outliers, we expect $d_{ij}-d_{lk}$ to be approximately equal to $y_c$, therefore we have $\gamma_c=0$. On the contrary, when the prediction of $d_{ij}-d_{lk}$ differs greatly from $y_c$, we can explain $\boldsymbol{e}^{\mathcal{U}}_c$ as the outliers and compensate for the discrepancy between the prediction and the annotation with a nonzero value of $\gamma_c$. The only prior knowledge we have on $\boldsymbol{\gamma}$ is that it is a sparse variable, \textit{i.e.,} in most cases $\gamma_c = 0$. 

Thanks to the low computational complexity and convenience in optimization, we will consider a linear model. We replace the embedding matrix $\boldsymbol{X}$ with the Gram matrix $\boldsymbol{G}=\boldsymbol{X}^{\top}\boldsymbol{X}=\{g_{ij}\}$ in \eqref{eq:linear_model} as
\begin{equation}
	\label{eq:linear_model_Gram}
	\begin{aligned}
	y_c 
= g_{ii}-2g_{ij}+g_{jj}-g_{ll}+2g_{lk}-g_{kk}+\gamma_c+\varepsilon_c.
	\end{aligned}
\end{equation}
Next we define the gradient operator (finite difference operator) on graph $\mathcal{G}$ which maps a dissimilarity function on the vertices $d:\mathcal{V}\rightarrow\mathbb{R}$ to an edge flow grad $\nabla d:\mathcal{V}\times\mathcal{V}\rightarrow\mathbb{R}$ such that $\nabla d\ (v_{ij}, v_{lk})=d_{ij}-d_{lk}$. For the whole graph $\mathcal{G}$, \eqref{eq:linear_model_Gram} can be re-written in the matrix form as
\begin{equation}
	\label{eq:linear_model_matrix_form}
	\boldsymbol{y} = \boldsymbol{Z}\odot\boldsymbol{G} + \boldsymbol{\gamma} + \boldsymbol{\varepsilon}, 
\end{equation}
and we define 
\begin{equation*}
	\boldsymbol{Z}\odot\boldsymbol{G}=\boldsymbol{Zg}=\boldsymbol{Z}\cdot\textit{vec}(\boldsymbol{G}),
\end{equation*}
where $\boldsymbol{Z}=\nabla\in\mathbb{R}^{|\mathcal{E}|\times n^2}$ is the design matrix, $\boldsymbol{y},\ \boldsymbol{\gamma}$ and $\boldsymbol{\varepsilon}$ are vectors in $\mathbb{R}^{|\mathcal{E}|}$. It is easy to see that \eqref{eq:linear_model_matrix_form} is a linear model.

In order to estimate the $|\mathcal{E}| + n^2$ unknown parameters ($|\mathcal{E}|$ for $\boldsymbol{\gamma}$ and $n^2$ for $\boldsymbol{G}$), we aim to minimize the discrepancy between the annotation $\boldsymbol{y}$ and the prediction $\boldsymbol{Z}\odot\boldsymbol{G} + \boldsymbol{\gamma}$, as well as holding the outlier indicator $\boldsymbol{\gamma}$ sparse. In addition, the estimated $\boldsymbol{G}$ should be a positive semi-definite matrix and its rank would be no more than $p\ll n$. Note that $\boldsymbol{y}$ only contains information about direction, but not how many votes received by each multiple edge $\boldsymbol{e}^{\mathcal{U}}_c$. The discrepancy thus needs to be weighted by the number of votes, represented by the edge weight vector $\boldsymbol{w}$. To that end, we put a weighted $\ell_2$-loss on the discrepancy, a sparsity enhancing penalty on $\boldsymbol{\gamma}$ as well as semi-definite positive and rank constraint on $\boldsymbol{G}$. It gives us the following optimization problem:
\begin{equation}
	\label{opt:roe}
	\begin{aligned}
		& &\underset{\boldsymbol{G},\ \boldsymbol{\gamma}}{\textit{minimize}}&\ \ \mathcal{L}_{\boldsymbol{w}}(\boldsymbol{G},\boldsymbol{\gamma}) +\lambda\|\boldsymbol{\gamma}\|_{1,\boldsymbol{w}}\\
		& &\textit{subject to}& \ \  \textbf{\textit{rank}}(\boldsymbol{G})= p,\ \boldsymbol{G}\succeq 0,
	\end{aligned}
\end{equation}
where
\begin{equation}
	\label{eq:weight_l2_loss}
	\begin{aligned}
		& \mathcal{L}_{\boldsymbol{w}}(\boldsymbol{G},\boldsymbol{\gamma}) &=&\ \ \frac{1}{2}\|\boldsymbol{y} - \boldsymbol{Z}\odot\boldsymbol{G} - \boldsymbol{\gamma}\|^2_{2,\boldsymbol{w}}\\
		& &=&\ \ \frac{1}{2}\underset{\boldsymbol{e}^{\mathcal{U}}_c\in\mathcal{E}}{\sum}w^2_c(y_c-\gamma_c-d_{ij}+d_{lk})^2\\
		& &=&\ \ \frac{1}{2}\|\boldsymbol{Wy} - (\boldsymbol{WZ})\odot\boldsymbol{G} - \boldsymbol{W}\boldsymbol{\gamma}\|^2_{2},
	\end{aligned}
\end{equation}
and 
\begin{equation}
	\label{eq:weight_l1_reg}
	\|\boldsymbol{\gamma}\|_{1,\boldsymbol{w}}=\underset{\boldsymbol{e}^{\mathcal{U}}_c\in\mathcal{E}}{\sum}w_c|\gamma_c|=\|\boldsymbol{W}\boldsymbol{\gamma}\|_1,
\end{equation}
where $\boldsymbol{W}=\textit{Diag}(\boldsymbol{w})$ is the diagonal matrix of $\boldsymbol{w}$. Solving \eqref{opt:roe}, our framework identifies outliers globally by integrating all relative similarity measurements together, and obtains the embedding matrix $\boldsymbol{X}$ via eigen-decomposition of $\boldsymbol{G}$. The noise term $\boldsymbol{\varepsilon}$ has been ignored in \eqref{opt:roe} because the discrepancy is mainly caused by outliers due to their larger magnitude. 

\subsection{Optimization}
Let $\boldsymbol{y}_{\boldsymbol{w}}=\boldsymbol{Wy}$ and $\boldsymbol{g}=\textit{vec}(\boldsymbol{G})$, we do variable substitution in the weighted $\ell_2$-loss \eqref{eq:weight_l2_loss} as
\begin{equation}
	\label{eq:l2_loss}
	\begin{aligned}
		& & &\ \ \mathcal{L}_{\boldsymbol{w}}(\boldsymbol{G},\boldsymbol{\gamma})\\
		& &=&\ \ \frac{1}{2}\ \|\boldsymbol{y}_{\boldsymbol{w}} - \boldsymbol{WZ \cdot g} - \boldsymbol{W\gamma}\|^2_2\\
		& &=& \ \ \frac{1}{2}\left\|\boldsymbol{y}_{\boldsymbol{w}}-
		\begin{bmatrix}
 			\boldsymbol{WZ} & \\ 
 			& \boldsymbol{W}
		\end{bmatrix}
		\begin{pmatrix}
 			\boldsymbol{g} \\ 
 			\boldsymbol{\gamma}
		\end{pmatrix}
		\right\|^2_2\\
		& &=&\ \ \frac{1}{2}\ \|\boldsymbol{y}_{\boldsymbol{w}} - \boldsymbol{A\cdot\beta}\|^2_2:=f(\boldsymbol{\beta}),
	\end{aligned}
\end{equation}
and the sparsity enhancing penalty \eqref{eq:weight_l1_reg} can be re-written as
\begin{equation}
	\label{eq:l1_reg}
	\begin{aligned}
		\lambda\|\boldsymbol{\gamma}\|_{1,\boldsymbol{w}} = 
		\lambda\left\|
		\begin{bmatrix}
 			\boldsymbol{0} & \\ 
 			& \boldsymbol{W}
		\end{bmatrix}
		\begin{pmatrix}
 			\boldsymbol{g} \\ 
 			\boldsymbol{\gamma}
		\end{pmatrix}
		\right\|_1=\lambda\|\boldsymbol{B\cdot\beta}\|_1:=g(\boldsymbol{\beta}).
	\end{aligned}
\end{equation}
With \eqref{eq:l2_loss}, \eqref{eq:l1_reg} and ignoring the constraints on $\boldsymbol{G}$, \eqref{opt:roe} is equivalent to a \textit{LASSO} formulation
\begin{equation}
	\label{opt:lasso}
 	\underset{\boldsymbol{\beta}}{\arg\min}\ F(\boldsymbol{\beta}) := f(\boldsymbol{\beta})+g(\boldsymbol{\beta}).
\end{equation}
Let $\mathcal{F}$ be the solution set of \eqref{opt:lasso} and suppose $\boldsymbol{G}^*$ is a optimal solution of \eqref{opt:roe}, it holds that $\boldsymbol{G}^*\in\mathcal{F}$. As a consequence, we come to a semi-definite programming with rank equality constraint
\begin{equation}
	\label{opt:sdp_rank}
	\begin{aligned}
		& &\textit{find}&\ \ \boldsymbol{G},\ \boldsymbol{\gamma}\\
		& &\textit{subject to}&\ \ \boldsymbol{G,\ \gamma}\in\mathcal{F},\ \boldsymbol{G}\succeq 0,\ \textbf{\textit{rank}}(\boldsymbol{G})= p.
	\end{aligned}
\end{equation}
Solving a SDP with rank equality constraints like \eqref{opt:sdp_rank} is notoriously difficult. It is proposed in \cite{dattorro2010convex} to solve this problem via iteratively solving the following two convex problems:
\begin{subequations}
	\begin{align}
		& \underset{\boldsymbol{G},\ \boldsymbol{\gamma}}{\textit{minimize}} \ \ \left\langle\boldsymbol{G},\boldsymbol{K}^*\right\rangle\label{opt:convex_iter_1}\\
		& \textit{subject to}\ \ \boldsymbol{G,\ \gamma}\in\mathcal{F},\ \boldsymbol{G}\succeq 0\nonumber,\\
		& \underset{\boldsymbol{G}}{\textit{minimize}} \ \ \left\langle\boldsymbol{G}^*,\boldsymbol{K}\right\rangle\label{opt:convex_iter_2}\\
		& \textit{subject to}\ \ \textbf{\textit{trace}}(\boldsymbol{K})=n-p,\ \boldsymbol{0}\preceq\boldsymbol{K}\preceq \boldsymbol{I}\nonumber,
	\end{align}
\end{subequations}
where $\left\langle\boldsymbol{G},\boldsymbol{K}\right\rangle=\textbf{\textit{trace}}(\boldsymbol{K}^\top\boldsymbol{G})$, $\boldsymbol{G}^*$ is an optimal solution of \eqref{opt:convex_iter_1} and $\boldsymbol{K}^*$ is an optimal solution of \eqref{opt:convex_iter_2}. However, this iteration of the convex problem sequence generally produces a solution of $\boldsymbol{G}$ which satisfies $\textit{rank}(\boldsymbol{G})\leq p$\footnote{Because this iterative scheme for constraining rank in semidefinite program is not a projection method, it can find a rank-$p$ solution $\boldsymbol{G}^*$ only if at least one exists in the feasible set of \eqref{opt:sdp_rank}. When a rank-$p$ feasible solution to \eqref{opt:sdp_rank} exists, it remains an open problem to state conditions under which $\left\langle\boldsymbol{G}^*,\boldsymbol{K}^*\right\rangle=\sum_{i=p+1}^{n}\sigma_i(\boldsymbol{G}^*)=0$ is achieved by iterative solution of \eqref{opt:convex_iter_1} and \eqref{opt:convex_iter_2}.}. Meanwhile, the nuclear-norm heuristic which replaces the rank equality constraint in \eqref{opt:sdp_rank} with nuclear norm regularization often recovers a minimum-rank solution of an SDP feasibility problem \cite{DBLP:journals/siamrev/RechtFP10}. Tuning the free parameter in these methods to generate a rank-$p$ solution is computational intensive. Rounding methods, \textit{e.g.} low-rank projection which finds the rank-$p$ matrix that is closest to the positive semi-definite solution in some norm, can also be adopted to solve \eqref{opt:sdp_rank}. But this method just provides the low-rank approximated solutions instead the exact solution. To obtain the rank-$p$ solution of $\boldsymbol{G}$ explicitly, we leverage the ran-reduction for semi-definite programming \cite{DBLP:journals/ftopt/LemonSY16}. 



First, we solve the following optimization problem
\begin{equation}
	\label{opt:sdp}
	\begin{aligned}
		& &\textit{find}&\ \ \boldsymbol{G},\ \boldsymbol{\gamma}\\
		& &\textit{subject to}&\ \ \boldsymbol{G,\ \gamma}\in\mathcal{F},\ \boldsymbol{G}\succeq 0.
	\end{aligned}
\end{equation}
For any $L > 0$, consider the following quadratic approximation of $F(\boldsymbol{\beta}) := f(\boldsymbol{\beta})+g(\boldsymbol{\beta})$ at a given point $\boldsymbol{\beta}_0$:
\begin{equation}
	\label{eq:quad_approx}
	\begin{aligned}
		& Q_L(\boldsymbol{\beta},\boldsymbol{\beta}_0) &=&\ \ f(\boldsymbol{\beta}_0)+\langle\boldsymbol{\beta}-\boldsymbol{\beta}_0,\nabla f(\boldsymbol{\beta}_0)\rangle\\
	& & &\ \ +\frac{L}{2}\|\boldsymbol{\beta}-\boldsymbol{\beta}_0\|^2+g(\boldsymbol{\beta}),		
	\end{aligned}
\end{equation}
which admits a unique minimizer
\begin{equation*}
	\label{eq:proximal_sol}
	\begin{aligned}
		& & &P_L(\boldsymbol{\beta})\\
		& &=&\ \ \underset{\boldsymbol{\beta}}{\arg\min}\left\{g(\boldsymbol{\beta})+\frac{L}{2}\left\|\boldsymbol{\beta}-\left(\boldsymbol{\beta}_0-\frac{1}{L}\nabla f(\boldsymbol{\beta}_0)\right)\right\|^2\right\}.
	\end{aligned}
\end{equation*}
Note that the regularization $g(\boldsymbol{\beta})$ is a constant in term of $\boldsymbol{G}$, we introduce projected gradient descent to find a semi-definite positive solution and the iterative scheme is 
\begin{equation}
	\label{eq:solve_G}
	\boldsymbol{G}_{t+1} = \Pi_{\mathcal{S}^n_+}\left(\boldsymbol{G}_t-\frac{1}{L}\ \nabla_{\boldsymbol{G}}\ f(\boldsymbol{\beta}_t)\right),
\end{equation}
where $\Pi_{\mathcal{S}^n_+}$ is the semi-definite positive projection of a matrix in the Frobenius norm. Since the $\ell_1$ norm is separable, the computation of $\boldsymbol{\gamma}$ reduces to solving a one-dimensional minimization problem for each of its components,
\begin{equation}
	\label{eq:solve_gamma}
	\boldsymbol{\gamma}_{t+1} = \mathcal{T}_{\boldsymbol{\mu}}\left(\boldsymbol{\gamma}_t-\frac{1}{L}\ \nabla_{\boldsymbol{\gamma}}\ f(\boldsymbol{\beta}_t)\right), 
\end{equation}
where $\mathcal{T}_{\boldsymbol{\mu}}:\mathbb{R}^{|\mathcal{E}|}\rightarrow\mathbb{R}^{|\mathcal{E}|}$ is the shrinkage operator 
\begin{equation*}
	[\mathcal{T}_{\boldsymbol{\mu}}(\boldsymbol{\gamma})]_i=\max(|\gamma_i|-\mu_i, 0)\cdot\textbf{\textit{sign}}(\gamma_i),
\end{equation*}
and $\boldsymbol{\mu}=\frac{\lambda}{L}\boldsymbol{w}$.

Suppose $\boldsymbol{\beta}^+$ is  a solution of \eqref{opt:sdp}
\begin{equation*}
	\boldsymbol{\beta}^+ = [\boldsymbol{G}^+,\ \boldsymbol{\gamma}^+],\ \boldsymbol{G}^+\in\mathcal{F}\cap\mathbb{S}^n_+,
\end{equation*}
where  $\mathbb{S}^n_+$ is the \textit{PSD} cone of $n$-dimensional matrix. The main step of rank reduction is finding a new solution $\boldsymbol{G}^*$ which satisfies $\textbf{\textit{rank}}(\boldsymbol{G}^*)<\textbf{\textit{rank}}(\boldsymbol{G}^+)$. Here we assume that $\textbf{\textit{null}}(\boldsymbol{G}^*)\subset\textbf{\textit{null}}(\boldsymbol{G}^+)$. Since $\boldsymbol{G}^+\in\mathbb{S}^n_+$, there exists a matrix $\boldsymbol{U}\in\mathbb{R}^{n\times p}$ such that $\boldsymbol{G}^+=\boldsymbol{UU}^\top$. We hope $\boldsymbol{G}^*$ has the following update rule:
\begin{equation}
	\label{eq:rank_reduction}
	\begin{aligned}
		& \boldsymbol{G}^*&=&\ \ \boldsymbol{G}^+ +\alpha\ \boldsymbol{U\Delta U}^\top\\
		& &=&\ \ \boldsymbol{U}(\boldsymbol{I}+\alpha\boldsymbol{\Delta})\boldsymbol{U}^\top, 
	\end{aligned}
\end{equation}
where $\alpha<0$ is a step size and $\boldsymbol{\Delta}\in\mathbb{S}^p_+$ is the update direction. Here we reformulate \eqref{opt:sdp} as a standard SDP 
\begin{equation}
	\label{opt:sdp_s}
	\begin{aligned}
		& &\textit{minimize}&\ \ \|\boldsymbol{\gamma}\|_{1,\boldsymbol{w}}\\
		& &\textit{subject to}&\ \ \left\langle\boldsymbol{G}, \boldsymbol{A}_c\right\rangle+\gamma_c=y_c,\ \boldsymbol{e}^{\mathcal{U}}_c\in\mathcal{E},\\
		& & &\ \ \ \ \boldsymbol{G}\succeq 0,
	\end{aligned}
\end{equation}
where $\boldsymbol{A}_c\in\mathbb{R}^{n\times n}$ is a symmetric matrix which has zero entry everywhere except on the entries corresponding to $c=(i,j,l,k)$ which has the form
\begin{equation}
	\label{eq:quadratic}
	\boldsymbol{A}_c = \begin{blockarray}{crrrr}
	 &i&j&l&k\\
	\begin{block}{c(rrrr)}
	i&1&-1&0&0\\
	j&-1&1&0&0\\
	l&0&0&-1&1\\
	k&0&0&1&-1{}\\
	\end{block}
	\end{blockarray}.
\end{equation}
Due to $\boldsymbol{G}^+,\ \boldsymbol{\gamma}^+$ is a solution of \eqref{opt:sdp_s}, we need $\boldsymbol{G}^*$ satisfies the equality constraints
\begin{equation}
 	\left\langle\boldsymbol{G}^*, \boldsymbol{A}_c\right\rangle+\gamma^+_c=y_c,\ \boldsymbol{e}^{\mathcal{U}}_c\in\mathcal{E}.
 \end{equation}
 Substituting \eqref{eq:rank_reduction} into them and simplifying give the conditions 
\begin{equation}
	\label{eq:rank_reduc_condition}
	\left\langle\boldsymbol{\Delta},\ \boldsymbol{U}^\top\boldsymbol{A}_c\boldsymbol{U}\right\rangle = 0,\ \boldsymbol{e}^{\mathcal{U}}_c\in\mathcal{E}.
\end{equation}
For convenience we define the mapping $\mathcal{A}_{\boldsymbol{U}}:\mathbb{S}^p_+\rightarrow\mathbb{R}^{|\mathcal{E}|}$ such that
\begin{equation*}
	\mathcal{A}_{\boldsymbol{U}}(\boldsymbol{\Delta})=
	\begin{bmatrix}
		\left\langle\boldsymbol{\Delta},\ \boldsymbol{U}^\top\boldsymbol{A}_1\boldsymbol{U}\right\rangle\\ 
		\vdots \\ 
		\left\langle\boldsymbol{\Delta},\ \boldsymbol{U}^\top\boldsymbol{A}_{|\mathcal{E}|}\boldsymbol{U}\right\rangle
	\end{bmatrix}=
	\begin{bmatrix}
		\left\langle\boldsymbol{A}_1,\ \boldsymbol{U}\boldsymbol{\Delta}\boldsymbol{U}^\top\right\rangle\\ 
		\vdots \\ 
		\left\langle\boldsymbol{A}_{|\mathcal{E}|},\ \boldsymbol{U}\boldsymbol{\Delta}\boldsymbol{U}^\top\right\rangle
	\end{bmatrix}.
\end{equation*}
Then we can express the condition \eqref{eq:rank_reduc_condition} as 
\begin{equation*}
	\mathcal{A}_{\boldsymbol{U}}(\boldsymbol{\Delta})= 0,\ \text{ or }\ \boldsymbol{\Delta}\in\textbf{\textit{null}}(\mathcal{A}_{\boldsymbol{U}}).
\end{equation*}
 
Moreover, we choose $\alpha<0$ to make $\textbf{\textit{rank}}(\boldsymbol{G}^*)<\textbf{\textit{rank}}(\boldsymbol{G}^+)$ and keep $\boldsymbol{G}^*$ be a solution of \eqref{opt:sdp}. It means that
\begin{equation*}
	\boldsymbol{I}+\alpha\boldsymbol{\Delta}\in\mathbb{S}^n_+,
\end{equation*}
and $\boldsymbol{I}+\alpha\boldsymbol{\Delta}$ is singular. The process of rank reduction is given as Algorithm \ref{alg:rank_reduction}. The following theorem guarantees the existence of $\boldsymbol{G}^*$, the part of a solution of \eqref{opt:sdp} with 
\begin{equation*}
	\textbf{\textit{rank}}(\boldsymbol{G}^*)=p,\ \frac{p(p + 1)}{2} \leq |\mathcal{C}|,
\end{equation*}
where $|\mathcal{C}|$ is the number of linear equality constraints in \eqref{opt:sdp}. As the low-dimensional embedding requires that $p\ll n$ and the number of comparisons $|\mathcal{C}|$ is suggested to be $\boldsymbol{O}(pn\log n)$ \cite{NIPS2016_6554}, such a $\boldsymbol{G}^*$ could be solved efficiently via Algorithm \ref{alg:rank_reduction}.

\begin{theorem}{\cite{DBLP:journals/ftopt/LemonSY16}}
If \eqref{opt:sdp} is solvable, then it has a solution contains $\boldsymbol{G}^*$ with $\textbf{\textit{rank}}(\boldsymbol{G}^*)=p$ such that $p(p + 1)/2 \leq |\mathcal{C}|$. Moreover, Algorithm \ref{alg:rank_reduction} efficiently finds such $\boldsymbol{G}^*$.
\end{theorem}

\begin{algorithm}[thb!]
	\caption{$\textbf{\textit{rank-reduction}}(\boldsymbol{\beta}^+, p) $}
	\label{alg:rank_reduction}
	\KwIn{$\boldsymbol{\beta}^+=[\boldsymbol{G}^+, \boldsymbol{\gamma}^+]$ is a solution of \eqref{opt:sdp_s},\ \ \ \ \ \ \ \ \ \ \ \ \ \ \ \ \  \ \ \ \ \ \ $p$ is the embedding dimension.}
	\KwOut{$\boldsymbol{G}^*$, which satisfies $\textbf{\textit{rank}}(\boldsymbol{G}^*)=p$.}
	Initialize $\boldsymbol{G}^*=\boldsymbol{G}^+$\;
	\Repeat{$\textbf{\textit{null}}(\mathcal{A}_{\boldsymbol{U}})=\{0\}$}
	{
		$\boldsymbol{G}^*=\boldsymbol{UU}^{\top},\ \boldsymbol{U}\in\mathbb{R}^{n\times p}$\;
		find a nonzero $\boldsymbol{\Delta}\in \textbf{\textit{null}}(\mathcal{A}_{\boldsymbol{U}})$ (if possible)\;
		find a maximum-magnitude eigenvalue $\sigma_1$ of $\boldsymbol{\Delta}$\;
		update
		\begin{equation*}
			\boldsymbol{G}^* = \boldsymbol{U}\left(\boldsymbol{I}-\frac{1}{\sigma_1}\boldsymbol{\Delta}\right)\boldsymbol{U}^\top.
		\end{equation*}
	}
\end{algorithm}

At the end of this section, we summarize the whole optimization algorithm as Algorithm \ref{alg:FISTA_rank_reduction}. The reproducible code can be found here\footnote{\url{https://github.com/alphaprime/ROE}}.

\begin{algorithm}[thb!]
	\caption{FISTA with rank reduction for \eqref{opt:sdp_rank}}
	\label{alg:FISTA_rank_reduction}
	\KwIn{Comparison graph $\mathcal{G}$, the edge direction flag $\boldsymbol{y}$, multiple edge weight $\boldsymbol{w}$, the regularization parameter $\lambda$ and the embedding dimension $p$.}
	\KwOut{$\boldsymbol{G}^*$,\ $\boldsymbol{\gamma}^*$.}
	Initialize $\boldsymbol{G}_0\in\mathbb{R}^{n\times n}$, $\boldsymbol{\gamma}_0\in\mathbb{R}^{|\mathcal{E}|}$, $\boldsymbol{\beta}_0 = [\boldsymbol{G}_0, \boldsymbol{\gamma}_0]$; Set $L_0>0$, $\eta>1$, $t_1=1$, $k=1$, ${\boldsymbol{\beta}}^*_1 =\boldsymbol{\beta}_0$\;
	\While{not satisfies the stopping rules}
	{	
		Find the smallest nonnegative integer $i_k$ such that with $\bar{L} = \eta^{i_k}L_{k-1}$
		\begin{equation*}
			\label{eq:backtracking}
			F(P_{\bar{L}}({\boldsymbol{\beta}}^*_k))\leq Q_{\bar{L}}(P_{\bar{L}}({\boldsymbol{\beta}}^*_k),\ {\boldsymbol{\beta}}^*_k)
		\end{equation*}\\
		Set $L_k = \eta^{i_k}L_{k-1}$ and update
		\begin{equation*}
			\begin{aligned}
				& \boldsymbol{\beta}_k &=&\ \ \ P_{L_k}({\boldsymbol{\beta}}^*_k)\text{ via }\eqref{eq:solve_G}\text{ and }\eqref{eq:solve_gamma}\\
				& t_{k+1} &=&\ \ \ \frac{1+\sqrt{1+4t^2_k}}{2}\\
				& \tilde{\boldsymbol{\beta}}_{k+1} &=&\ \ \ \boldsymbol{\beta}_k+\frac{t_k-1}{t_{k+1}}\left(\boldsymbol{\beta}_k-\boldsymbol{\beta}_{k-1}\right)\\
				& {\boldsymbol{\beta}}^*_{k+1} &=&\ \ \ \textbf{\textit{rank-reduction}}(\tilde{\boldsymbol{\beta}}_{k+1}) \\
			\end{aligned}
		\end{equation*}\\
		$k\leftarrow k+1$\;
	}
\end{algorithm}

\section{Experiments}

{
	\begin{table*}[ht]
		\caption{Classification error results on the synthetic dataset.}
		\centering
		\label{tab:synthetic} 
		\begin{subtable}[c]{\columnwidth}
		  \centering
		  	\caption{Clean Data (without outlier)}
		    \begin{tabular}{lcccc}
			    \toprule
			    Methods & min   & median & max   & std \\
			    \midrule
			    GNMDS-$p$ & 0.2061 & 0.2434 & 0.2607 & 0.0194 \\
			    CKL-$p$ & 0.1879 & 0.1979 & 0.2088 & 0.0062 \\
			    STE-$p$ & 0.1829 & 0.1884 & 0.1930 & 0.0031 \\
			    Ours  & \textbf{0.0953} & \textbf{0.0961} & \textbf{0.1049} & \textbf{0.0024} \\
			    \bottomrule
		    \end{tabular}
			\label{subtab:synthetic:0}
		\end{subtable}
		\begin{subtable}[c]{\columnwidth}
		  \centering
			\caption{Contaminated Data (with $25\%$ outliers)}
			\begin{tabular}{lcccc}
				\toprule
			    Methods & min   & median & max   & std \\
			    \midrule
			    GNMDS-$p$ & 0.2373 & 0.2487 & 0.2586 & 0.0063 \\
			    CKL-$p$ & 0.1443 & 0.1632 & 0.1809 & 0.0103 \\
			    STE-$p$ & 0.1828 & 0.2035 & 0.2250 & 0.0127 \\
			    Ours  & \textbf{0.0960} & \textbf{0.1025} & \textbf{0.1070} & \textbf{0.0032} \\
			    \bottomrule
		    \end{tabular}
		  \label{subtab:synthetic:25}
		\end{subtable}
	\end{table*}	
}

{
	\begin{table*}[ht]
		\caption{Classification error results on the music artists dataset.}
		\label{tab:music}
		\centering
		\begin{subtable}[c]{\columnwidth}
		  \centering
		  \caption{Clean Data (without outlier)}
		    \begin{tabular}{lcccc}
			    \toprule
			    Methods & min   & median & max   & std \\
			    \midrule
			    GNMDS-$p$ & 0.2373 & 0.2506 & 0.2828 & 0.0101 \\
			    CKL-$p$ & \textbf{0.2239} & 0.2358 & 0.2813 & 0.0117 \\
			    STE-$p$ & 0.2278 & 0.2356 & 0.2763 & 0.0122 \\
			    Ours  & 0.2289 & \textbf{0.2312} & \textbf{0.2356} & \textbf{0.0016} \\
			    \bottomrule
		    \end{tabular}
		  \label{subtab:music:0}
		\end{subtable}
		\begin{subtable}[c]{\columnwidth}
		  \centering
			\caption{Contaminated Data (with $25\%$ outliers)}
		    \begin{tabular}{lcccc}
				\toprule
			    Methods & min   & median & max   & std \\
			    \midrule
			    GNMDS-$p$ & 0.2310 & 0.2460 & 0.2887 & 0.0121 \\
			    CKL-$p$ & 0.2099 & \textbf{0.2120} & 0.2505 & 0.0120 \\
			    STE-$p$ & \textbf{0.2027} & 0.2203 & 0.2579 & 0.0116 \\
			    Ours  & 0.2226 & 0.2287 & \textbf{0.2389} & \textbf{0.0047} \\
			    \bottomrule
		    \end{tabular}
		  \label{subtab:music:25}
		\end{subtable}
	\end{table*}	
}
\subsection{Simulation}

\noindent\textbf{Dataset. }The simulated dataset consists of $100$ points $\{\boldsymbol{x}_i\}_{i=1}^{100}\subset\mathbb{R}^{10}$, where $\boldsymbol{x}_i\sim\mathcal{N}(\boldsymbol{0}, \frac{1}{20}\boldsymbol{I})$, $\boldsymbol{I}\in \mathbb{R}^{10\times 10}$ is the identity matrix. The possible similarity triple-wise comparisons are generated based on the Euclidean distances between $\{\boldsymbol{x}_i\}$. We randomly sample $10000$ correct triplets as the basic training data and a validation set is build with the same number of triplets. The remains are served as the test set. 

\noindent\textbf{Settings. }The existing ordinal embedding methods adopt triple-wise comparisons as the constraints. The triple-wise comparison set $\mathcal{T}=\{(i,j,k)\}$ is the special case of quadruplet which means $l=i$ in $c=(i,j,l,k)\in\mathcal{C}$. The differences between the above triplets setting and the generalized formulation are two-fold, (i) the edge of triple-wise comparison graph $\mathcal{G}_{\mathcal{T}}=\{\mathcal{V}_{\mathcal{T}},\mathcal{E}_{\mathcal{T}}\}$ only links a pair of nodes which share the same item, such as $v_{ij}$ and $v_{ik}$, (ii) the equality constraints in \eqref{opt:sdp_s} could be modified as 
$
	\left\langle\boldsymbol{G}, \boldsymbol{A}_t\right\rangle+\gamma_t=y_t,\ \boldsymbol{e}^{\mathcal{U}}_t\in\mathcal{E}_{\mathcal{T}},
$
where $\boldsymbol{A}_t$ is a symmetric $n\times n$ matrix indicated by $t=(i,j,k)$ as 
\begin{equation}
	\boldsymbol{A}_t = \begin{blockarray}{crrr}
	 &i&j&k\\
	\begin{block}{c(rrr)}
	i&0&-1&1\\
	j&-1&1&0\\
	k&1&0&-1\\
	\end{block}
	\end{blockarray}\ \ .
\end{equation}
With these modifications, the proposed method employs triple-wise comparisons $\mathcal{T}$ to construct the comparison graph $\mathcal{G}_{\mathcal{T}}$ for fair competition.

The basic training comparisons are augmented to represent the votes received on a triplet. For each triplet $t$, there will be $s$ copies, $t_1, \dots, t_s,\ 15\leq s \leq 50$. Second, errors are then synthesized according to the different ratios: we assume that at most $q\%$ of all relative comparisons are not consistent with the ground-truth metric information and we change the position of $j$ and $k$ in each randomly chosen triplet, where $q$ ranges from $10\%$ to $25\%$. Thus the outliers of these comparisons could not be the minority of the edges between a pair of nodes. Specially, we conduct the experiments with the ``noiseless'' case where $q=0$ which indicates that the local outlier detection can detect all outliers in the augmented training comparisons. Actually, this setting is just an ideal case which would not be expected in real applications. At last, the comparison graph $\mathcal{G}_{\mathcal{T}}$ is constructed, where $\mathcal{T}$ is the contaminated training set.

\noindent\textbf{Evaluation Metrics. }We employ the classification error to evaluate the performance of various algorithms. The learned Gram matrix $\boldsymbol{G}$ can predict the direction of an unseen but possible edge in the graph. The percentage of wrong prediction,  which is not consistent with the ground-truth metric information, is the classification error of the learned embedding. Larger classification error means lower quality of the learned embedding.

\noindent\textbf{Competitors. }We compare the proposed algorithm with three well-known ordinal embedding methods: GNMDS \cite{agarwal2007generalized}, CKL \cite{tamuz2011adaptiive} and STE \cite{vandermaaten2012stochastic}. Note that we adopt the strategies proposed by \cite{NIPS2016_6554}, which performs projected gradient descent with line search. The learned matrix are projected onto the subspace spanned by the top $p$ eigenvalues at each iteration, \textit{i.e.} setting the smallest $n-p$ eigenvalues to $0$. We call the three new algorithms: GNMDS-$p$, CKL-$p$ and STE-$p$, correspondingly. The parameters of these competitors are tuned on the validation set.

\begin{figure}[ht]
	\centering
	\begin{subfigure}[b]{0.45\columnwidth}
		\centering
		\includegraphics[width = .9\textwidth]{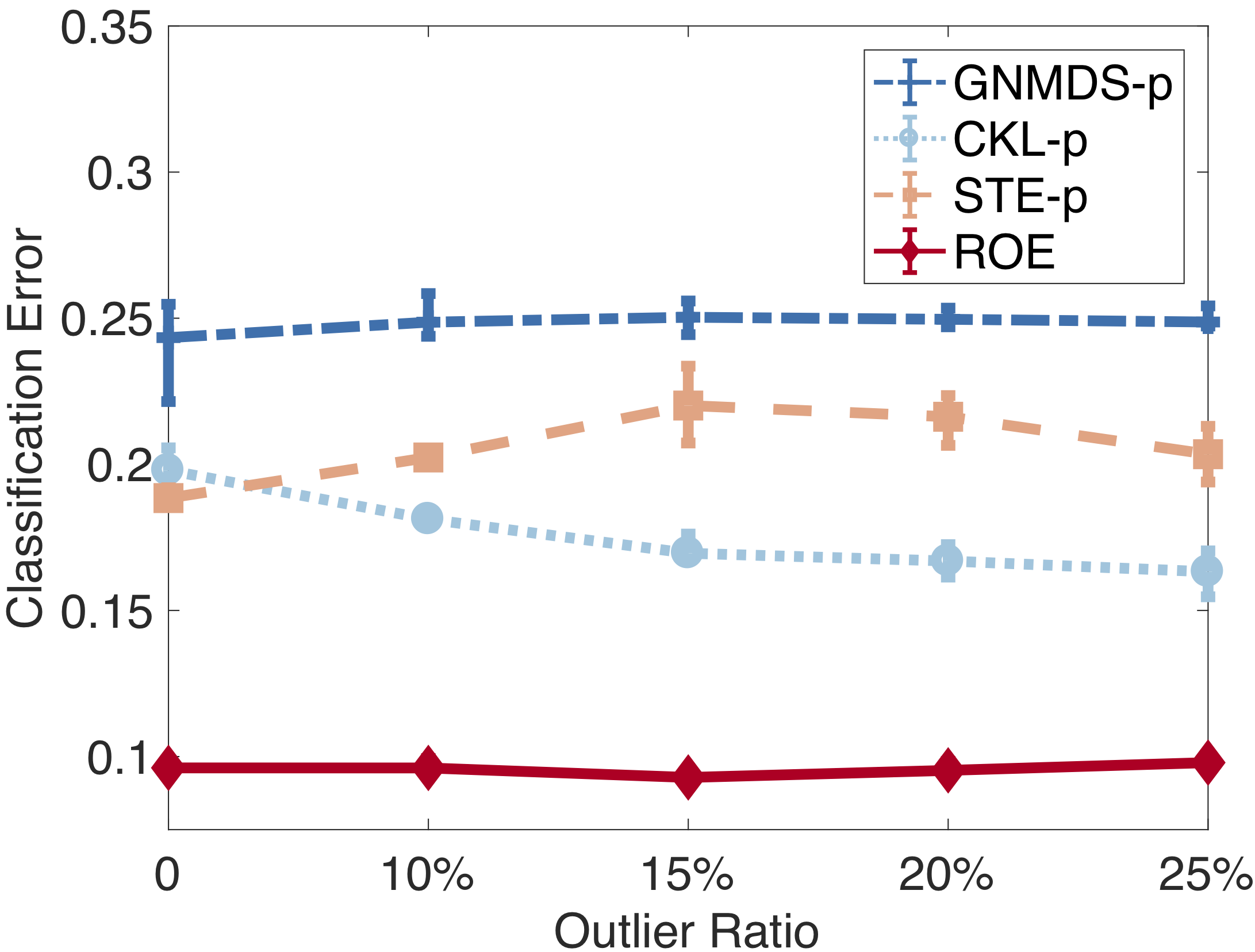}
		\caption{Comparative evaluation.}
		\label{fig:synthetic:error}
	\end{subfigure}
	\begin{subfigure}[b]{0.45\columnwidth}
		\centering
		\includegraphics[width = .9\textwidth]{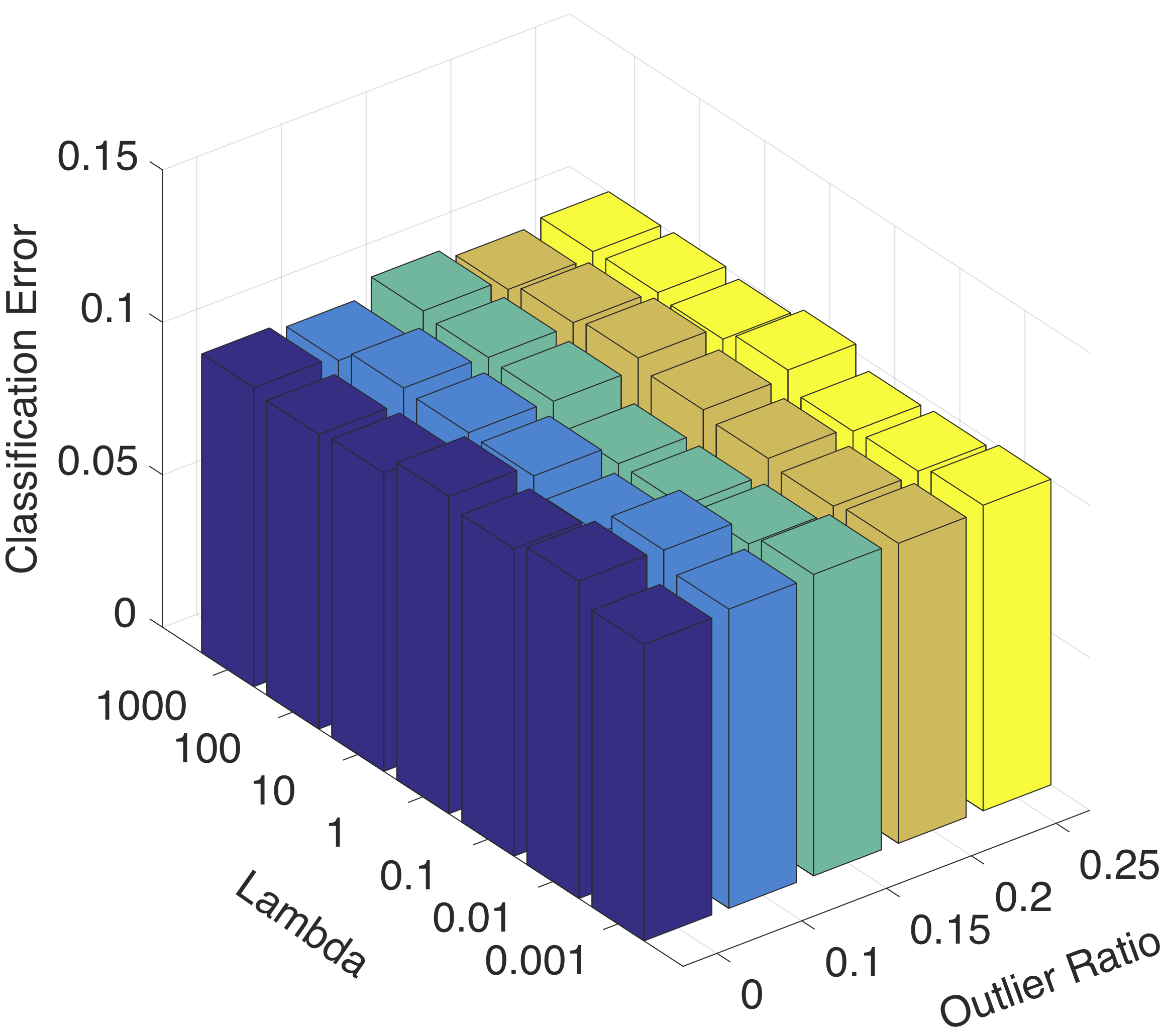}
		\caption{Parameter sensitivity.}
		\label{fig:synthetic:lambda}
	\end{subfigure}
	\label{fig:synthetic}
	\caption{The results on the synthetic data. (a) The classification performance comparative evaluation. Smaller classification error means better embedding result. The mean and standard deviation of each method over $20$ trials are shown in the plots. (b) The classification error with different choices of $\lambda$.}
\end{figure}

\noindent\textbf{Comparative Results. }The embedding performance of  various models are evaluated when different ratios of outliers are considered. The results are shown in Figure \ref{fig:synthetic:error} and Table \ref{tab:synthetic}. We provide more details in supplementary materials. It shows clearly that \textbf{\textit{ROE}} significantly outperforms the three alternatives for a wide range of noise density. This validates the effectiveness of \textbf{\textit{ROE}}. In particular, it can be observed that: (i) the improvement over the three competitors demonstrates the superior generalization ability of \textbf{\textit{ROE}} thanks to the unified framework rather than phased methodology. The traditional methods rely heavily on majority voting and maximum acyclic subgraph approximation as the preprocessing, but their models ignore the intrinsic inconsistencies between the noise comparisons and the ground-truth similarity relationship of $\boldsymbol{X}$. More important, these inconsistencies, especially the global ones, would not be conquered by the local outlier detection methods. Consequently, the three alternatives would suffer from the wrong training comparisons and the classification error would be amplified and accumulated. However, the proposed \textbf{\textit{ROE}} method incorporate the global outlier detection scheme with ordinal embedding. This unified framework not only benefits from the correct training samples which would be pruned by majority voting and maximum acyclic subgraph approximation, but also gets rid of the contamination from the outliers. (ii) Even under the ``noiseless'' situation, the proposed method also shows the superiority. This improvement comes from the exact rank-$p$ solution of Algorithm \ref{alg:rank_reduction} and the regression-based framework which aggregates all the votes on a comparison.

\noindent\textbf{Parameter Sensitivity Analysis. }To show the sensitivity of \textbf{\textit{ROE}} toward the free parameter $\lambda$ in \eqref{opt:roe} changes, we record the average classification error over $20$ runs for the synthetic datasets with different $\lambda$. The corresponding result are shown in Figure \ref{fig:synthetic:lambda}. We find that the performance of \textbf{\textit{ROE}} keeps relatively stable overall in a wide range, from $10^{-3}$ to $10^3$.

\begin{figure}[ht]
	\centering
	\begin{subfigure}[b]{0.45\columnwidth}
		\centering
		\includegraphics[width = .9\textwidth]{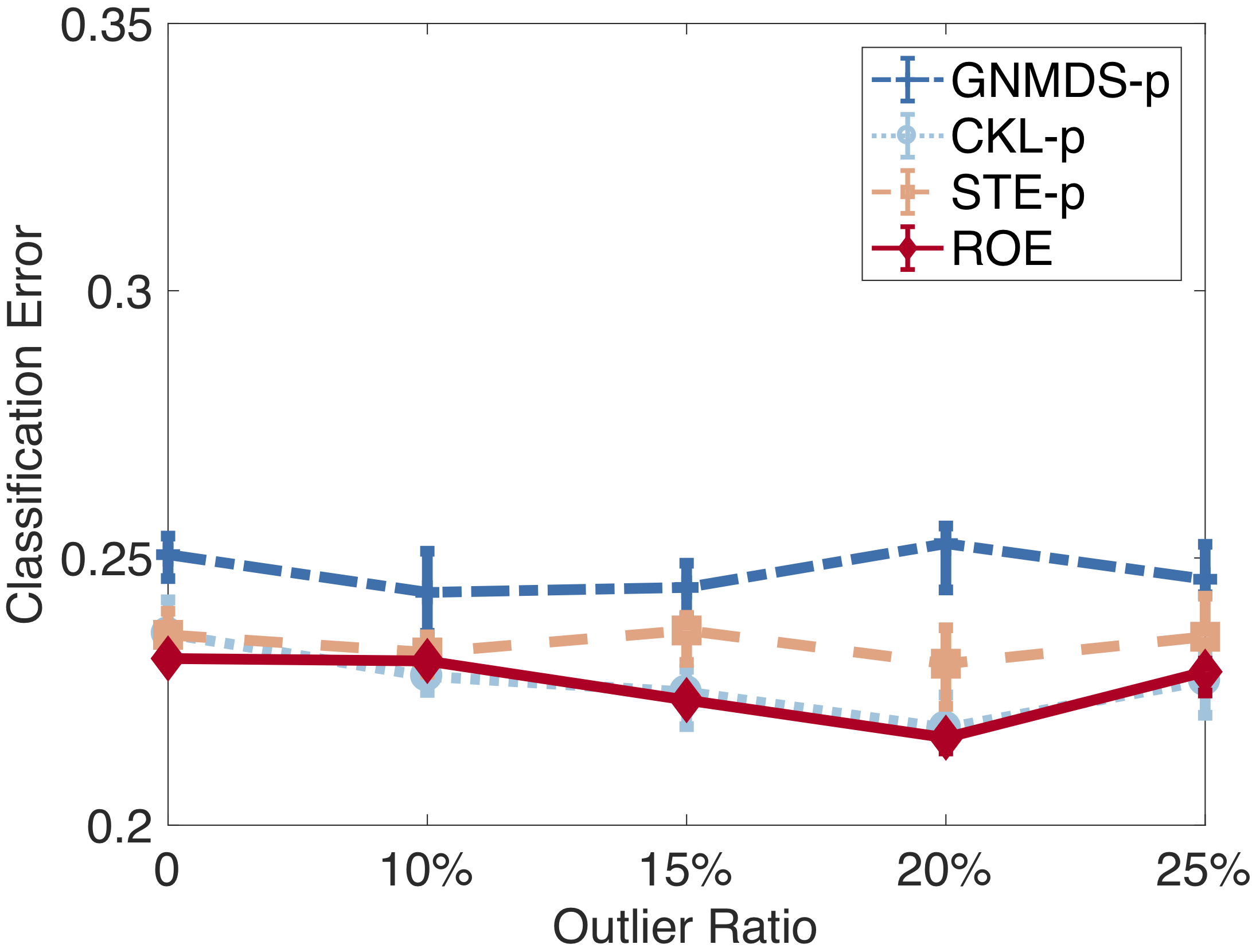}
		\caption{Comparative evaluation.}
		\label{fig:music:error}
	\end{subfigure}
	\begin{subfigure}[b]{0.45\columnwidth}
		\centering
		\includegraphics[width = .9\textwidth]{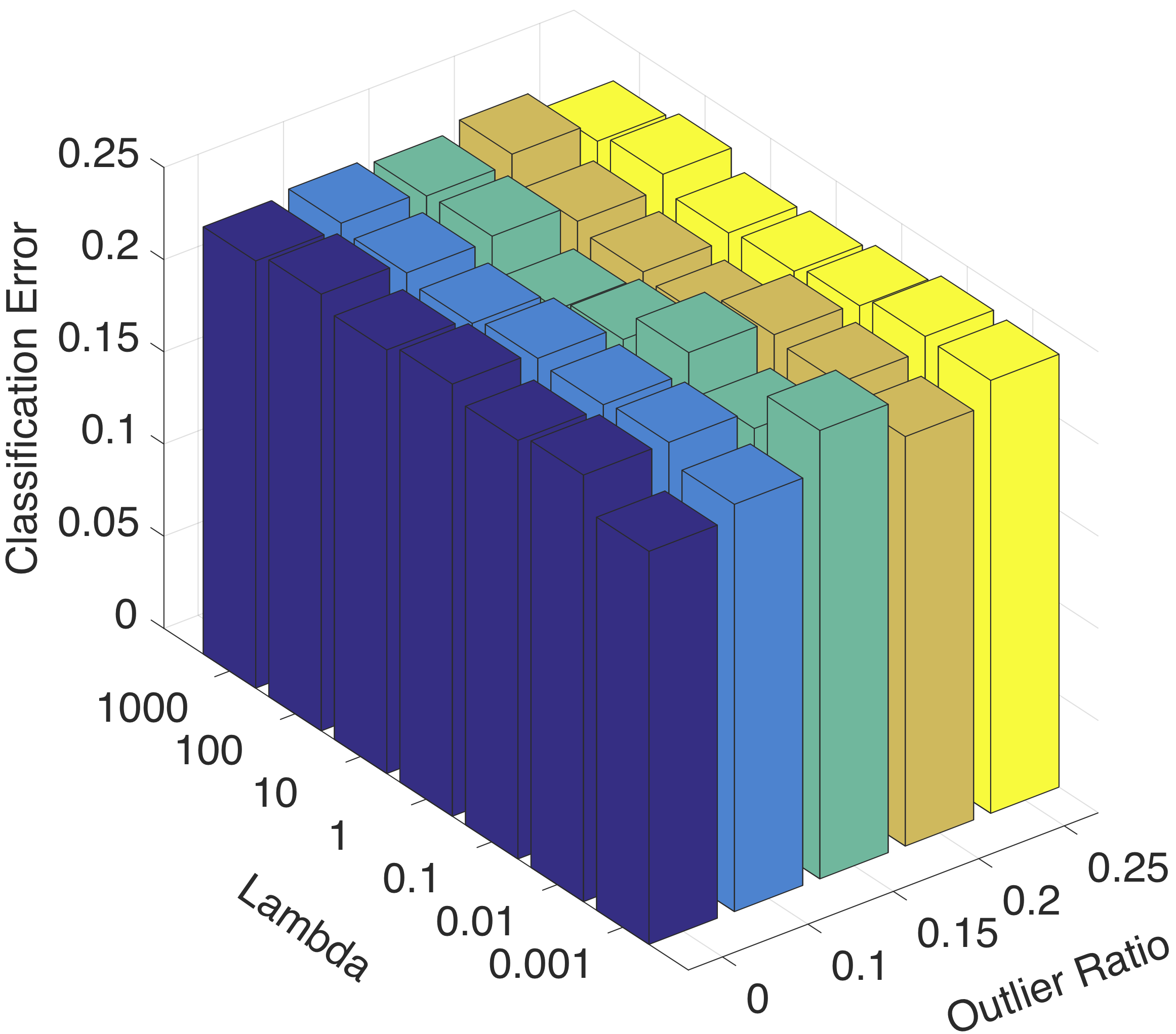}
		\caption{Parameter analysis.}
		\label{fig:music:lambda}
	\end{subfigure}
	\label{fig:music}
	\caption{The results on the music data. (a) The classification comparative evaluation. Smaller classification error means better embedding result. The mean and standard deviation of each method over $20$ trials are shown in the plots. (b) The sensitive analysis with different choices of $\lambda$.}
\end{figure}

\subsection{Music Artists Similarity}

\noindent\textbf{Dataset. }The music artist data is collected by \cite{ellis2002quest} via a web-based survey in which $1,032$ users provided triple-wise comparisons on the similarity of $412$ music artists. The traditional methods like \cite{vandermaaten2012stochastic} adopt the majority voting and the maximum acyclic subgraph approximation to prune the inconsistence comparisons. Therefore, a much smaller version\footnote{\url{https://lvdmaaten.github.io/ste/Stochastic_Triplet_Embedding.html}}, which has only $9,107$ triplets for $n=400$ artists, is employed by the existing methods. For fair comparison, we evaluate \textbf{\textit{ROE}} on this subset. The size of training samples is $5,000$ and the validation set contains $2,000$ triplets. The rest of triplets are treated as test set. The desired dimension of embedding is $d=9$ as these music artists can be classified by genre into $9$ categories. It's worth noting that these $9,107$ triplets still include outliers duo the inappropriate preprocessing. Accordingly, the evaluation on this data verifies that the local outlier detection and maximum acyclic subgraph approximation should be eliminated. 

\noindent\textbf{Comparative Results. } Without the ground truth of music artists similarity values, different models were evaluated
indirectly via classification accuracy based on the noise test comparisons in Figure \ref{fig:music:error}. The following observations can be made: (i) the wrong data would damage the performance of ordinal embedding methods. The outliers in training data cause the three alternatives to generate the sub-optimal solutions as they can't prune the outliers without preprocessing. The global outlier detection is more accurate but the wrong comparisons in validation set and test set would lead the evaluation metric of \textbf{\textit{ROE}} to be higher than these competitors. This phenomenon proves the effectiveness of \textbf{\textit{ROE}} from the opposite side. (ii) Benefit from the global outlier detection ability, the variance of \textbf{\textit{ROE}} is much smaller than these three alternatives because \textbf{\textit{ROE}} can prune the outliers and keep the result more stable. 

\section{Conclusions}
In this paper we introduce a novel unified robust framework to construct the representation of items in the Euclidean space $\mathbb{R}^p$ with contaminated comparisons. The key advantage of our method over the existing approaches is that our model infers the embedding and detects the outliers jointly by minimizing a global ranking inconsistence cost. It can be formulated as a partial penalized LASSO optimization problem. Efficient algorithm is proposed to obtain a positive semi-definite solution which satisfies the rank equality constraint. Experimental studies are conducted with both synthetic and real-world data. Our results suggest that the local outlier detection is not a reliable tool for ordinal embedding with contaminated comparisons.

\section*{Acknowledgment}
The research of Ke Ma and Xiaochun Cao is supported by the National Key R\&D Program of China (Grant No. 2016YFB0800603), the Key Program of the Chinese Academy of Sciences (No. QYZDB-SSW-JSC003) and the National Natural Science Foundation of China (No.U1636214, U1605252, 61733007). The research of Qianqian Xu is supported in part by the National Natural Science Foundation of China (No.61672514, 61390514, 61572042), the Beijing Natural Science Foundation (4182079), the Youth Innovation Promotion Association CAS, and the CCF-Tencent Open Research Fund.

\bibliographystyle{aaai}
\bibliography{aaai19}

\end{document}